\documentclass[12pt]{article}

\usepackage{sbc-template}
\usepackage{graphicx,url}
\usepackage[utf8]{inputenc}
\usepackage[english]{babel}
\usepackage{acronym}
\usepackage{xcolor}
\usepackage{multirow}
\usepackage{caption}
\usepackage{subcaption}
     
\title{On Enhancing Network Throughput using Reinforcement Learning in Sliced Testbeds}

\author{Daniel Pereira Monteiro\inst{1}, Lucas Nardelli de Freitas Botelho Saar\inst{1}, \\Larissa Ferreira {Rodrigues Moreira}\inst{1,2}, Rodrigo Moreira\inst{1}}

\address{Institute of Exact and Technological Sciences -- Federal University of Viçosa 
  (UFV)\\
  Rio Paranaíba -- MG -- Brazil.
\nextinstitute
  Faculty of Computing (FACOM) -- Federal University of Uberlândia (UFU)\\
  Uberlândia -- MG -- Brazil.
\email{\{daniel.p.monteiro, lucas.saar, larissa.f.rodrigues, rodrigo\}@ufv.br} \email{larissarodrigues@ufu.br}
}

\begin{document} 
\acrodef{3GPP}{3rd Generation Partnership Project}
\acrodef{AI}{Artificial Intelligence}
\acrodef{B5G}{Beyond Fifth Generation}

\acrodef{CUBIC}{Conjunctive Using BIC (Binary Increase Congestion Control)}
\acrodef{cwnd}{Congestion Window}
\acrodef{DoS}{Denial of Service}
\acrodef{DDoS}{Distributed Denial of Service}
\acrodef{DNN}{Deep Neural Network}
\acrodef{DRL}{Deep Reinforcement Learning}
\acrodef{DT}{Decision Tree}
\acrodef{DNN}{Deep Neural Network}
\acrodef{DMP} {Deep Multilayer Perceptron}
\acrodef{DQN}{Deep Q-Learning}
\acrodef{eMBB}{Enhanced Mobile Broadband}
\acrodef{ETSI}{European Telecommunications Standards Institute}
\acrodef{FIBRE}{Future Internet Brazilian Environment for Experimentation}
\acrodef{FTP}{File Transfer Protocol}
\acrodef{Flat}{Flat Neural Network}
\acrodef{GNN}{Graph Neural Networks}
\acrodef{HTM}{Hierarchical Temporal Memory}

\acrodef{IAM}{Identity And Access Management}
\acrodef{IID}{Informally, Identically Distributed}
\acrodef{IoE}{Internet of Everything}
\acrodef{IoT}{Internet of Things}
\acrodef{KNN}{K-Nearest Neighbors}
\acrodef{LSTM}{Long Short-Term Memory}
\acrodef{MPTCP}{Multipath Transmission Control Protocol}
\acrodef{M2M}{Machine to Machine}
\acrodef{MAE}{Mean Absolute Error}
\acrodef{ML}{Machine Learning}
\acrodef{MOS}{Mean Opinion Score}
\acrodef{MAPE}{Mean Absolute Percentage Error}
\acrodef{MSE}{Mean Squared Error}
\acrodef{mMTC}{Massive Machine Type Communications}
\acrodef{MFA}{Multi-factor Authentication}
\acrodef{MQTT}{Message Queuing Telemetry Transport}

\acrodef{DNN}{Deep Neural Network}
\acrodef{NN}{Neural Network}
\acrodef{NNs}{Neural Networks}
\acrodef{NS3}{Network Simulator 3}
\acrodef{OSM}{Open Source MANO}
\acrodef{QL}{Q-learning}
\acrodef{QoE}{Quality of experience}
\acrodef{QoS}{Quality of Service}
\acrodef{RAM}{Random-Access Memory}
\acrodef{RF}{Random Forest}
\acrodef{RL}{Reinforcement Learning}
\acrodef{RMSE}{Root Mean Square Error}
\acrodef{RNN}{Recurrent Neural Network}
\acrodef{Reno}{Regular NewReno}
\acrodef{RTT}{Round Trip Time}
\acrodef{SDN}{Software-Defined Networking}
\acrodef{SFI2}{Slicing Future Internet Infrastructures}
\acrodef{SLA}{Service-Level Agreement}
\acrodef{SON}{Self-Organizing Network}

\acrodef{TCP}{Transmission Control Protocol}
\acrodef{VoD}{Video on Demand}
\acrodef{VR}{Virtual Reality}
\acrodef{V2X}{Vehicle-to-Everything}

\maketitle

\begin{abstract}

Novel applications demand high throughput, low latency, and high reliability connectivity and still pose significant challenges to slicing orchestration architectures. The literature explores network slicing techniques that employ canonical methods, artificial intelligence, and combinatorial optimization to address errors and ensure throughput for network slice data plane. This paper introduces the \ac{eMBB}-Agent as a new approach that uses Reinforcement Learning (RL) in a vertical application to enhance network slicing throughput to fit Service-Level Agreements (SLAs). The eMBB-Agent analyzes application transmission variables and proposes actions within a discrete space to adjust the reception window using a Deep Q-Network (DQN). This paper also presents experimental results that examine the impact of factors such as the channel error rate, DQN model layers, and learning rate on model convergence and achieved throughput, providing insights on embedding intelligence in network slicing.
  
\end{abstract}

\section{Introduction}\label{sec:introduction}

Disruptive applications, such as 8K video streaming, Virtual Reality (VR), and Augmented Reality (AR), had led to an increased demand for high network throughput~\cite{Khan2022}. Additionally, other application families, including remote surgery, smart factories, and autonomous vehicles, require low-latency and high-reliability connectivity~\cite{Aripin2023}. Ensuring the compatibility between these conflicting requirements within a physical network is a significant challenge for both management and resource orchestration~\cite{Khan2022}. To address this, various advances have been made in network slicing, virtualization, programmability, security, and \ac{AI}, especially in mainstream mobile networks, such as beamforming and energy-aware solutions~\cite{Khan2022, Moreira2023, Brilhante2023}. 

The literature has explored approaches for performing network slicing that can handle errors in the underlying channel while ensuring throughput using canonical techniques such as artificial intelligence, and combinatorial optimization~\cite{Ojijo2020, Liu2023}. Some of these techniques involve intervention in the link~\cite{moreira2021_XDP}, while others involve intervention in the communicating entity, such as those based on \ac{TCP}~\cite{Li2019, Sadia2022} Congestion Control. The proposed control in the communicating entities aims to reduce the reception window in the event of packet loss, thereby reducing the amount of traffic in the network. However, this approach is not sufficiently flexible to incorporate intelligent mechanisms seamlessly.

In this paper, we propose and evaluate the \texttt{\ac{eMBB} Agent}, which is based on \ac{RL} coupled with a vertical application to improve the throughput of network slicing to guarantee \ac{SLA}. Functionally, \texttt{\ac{eMBB}-Agent} analyzes the vertical application variables and proposes actions within a discrete space to increase or decrease the reception window. Subsequently, the \texttt{\ac{eMBB}-Agent} verifies the effectiveness of its actions through an \ac{DQN}. Experimentally, we verified how factors such as the channel error rate, number of layers in the \ac{DQN} model, and learning rate impact the model convergence and achieved throughput.

The remainder of this paper is organized as follows. In Section~\ref{sec:related_work}, we contextualize the related work on intelligent throughput enhancement. The proposed experimental method is presented in detail in Section~\ref{sec:evaluation_method}, followed by a description of the experimental setup and results in Section~\ref{sec:results_and_discussion}. Section~\ref{sec:concluding_remark} discusses concluding remarks and future research directions.

\section{Related Work}\label{sec:related_work}

Recent efforts, such as \cite{Zhang2019}, have sought to improve the \textit{\ac{MPTCP} } protocol through reinforcement learning techniques. Using asynchronous training, this study allows parallel execution of packet scheduling, data collection, and neural network training. The goal was to optimize scheduling in real time by employing an asynchronous algorithm for neural training.

The work proposed by \cite{Li_Wenzhong2019} aimed to improve network efficiency using the \textit{SmartCC} algorithm. This algorithm employs reinforcement learning techniques to improve the congestion window management. \textit{SmartCC} uses an asynchronous reinforcement learning mechanism to acquire a set of congestion rules. While \cite{Tang2019} presented a traffic prediction algorithm based on deep learning. This algorithm aims to anticipate the workload and network congestion. After the prediction, partial channel allocation based on deep learning is performed to prevent possible congestion by assigning appropriate channels.

The study carried out by \cite{beig2018mptcp} examined mobile users using the \ac{MPTCP} protocol, with the aim of optimizing congestion control in heterogeneous networks. This study proposes an algorithm based on \textit{\ac{QL}} to improve \textit{throughput}, with the aim of maximizing throughput. \cite{Vieira_Garcez_2011} developed a mathematical expression to calculate the probability of data loss on the servers. This expression is used to condition the estimation of the probability of data loss in analog servers that have a finite \textit{buffer} and receive time-dependent multifactor flows.

Table \ref{tab:related_work} aims to clarify the standards adopted by related studies in relation to the metrics and technologies used. The \textit{throughput} metric is frequently used as an evaluation criterion in several studies. Another observed constant is its use as a search variable for designing experiments.

\begin{table}[htbp]
\caption{Prior works aimed to enhance network conditions using \ac{AI}.}
\label{tab:related_work}

\centering

\resizebox{\textwidth}{!}{%
\begin{tabular}{ccccc}
\hline
\textbf{Approach} & \textbf{Evaluation Metric} & \textbf{Search Variable} & \textbf{\ac{AI}} & \textbf{Evaluation Testbed} \\ \hline
\cite{Tang2019} & Throughput & \ac{SDN} Controller & Reinforcement Learning & API C++/
WILL \\
\cite{beig2018mptcp} & Throughput & Throughput & Q-Learning & NS3 \\
\cite{Vieira_Garcez_2011}& Probability, buffer & Buffer & Does Not Use & Own \\
\cite{Zhang2019} & Goodput, Delay, Download & MinRTT, Round-Robin & \ac{DRL} & Own \\
\cite{Li_Wenzhong2019} & ACK, \ac{RTT} & Throughput, \ac{RTT}, Jitter & \ac{RL} & Own \\ \textbf{Our Approach} & Throughput, and \ac{RTT} & Congestion Window & \ac{DQN} & NS3 on Fabric \\ \hline

\end{tabular}%
}
\end{table}

\section{Evaluation Method}\label{sec:evaluation_method}

\begin{figure}[htbp]
  \centering
  \includegraphics[width=0.53\textwidth]{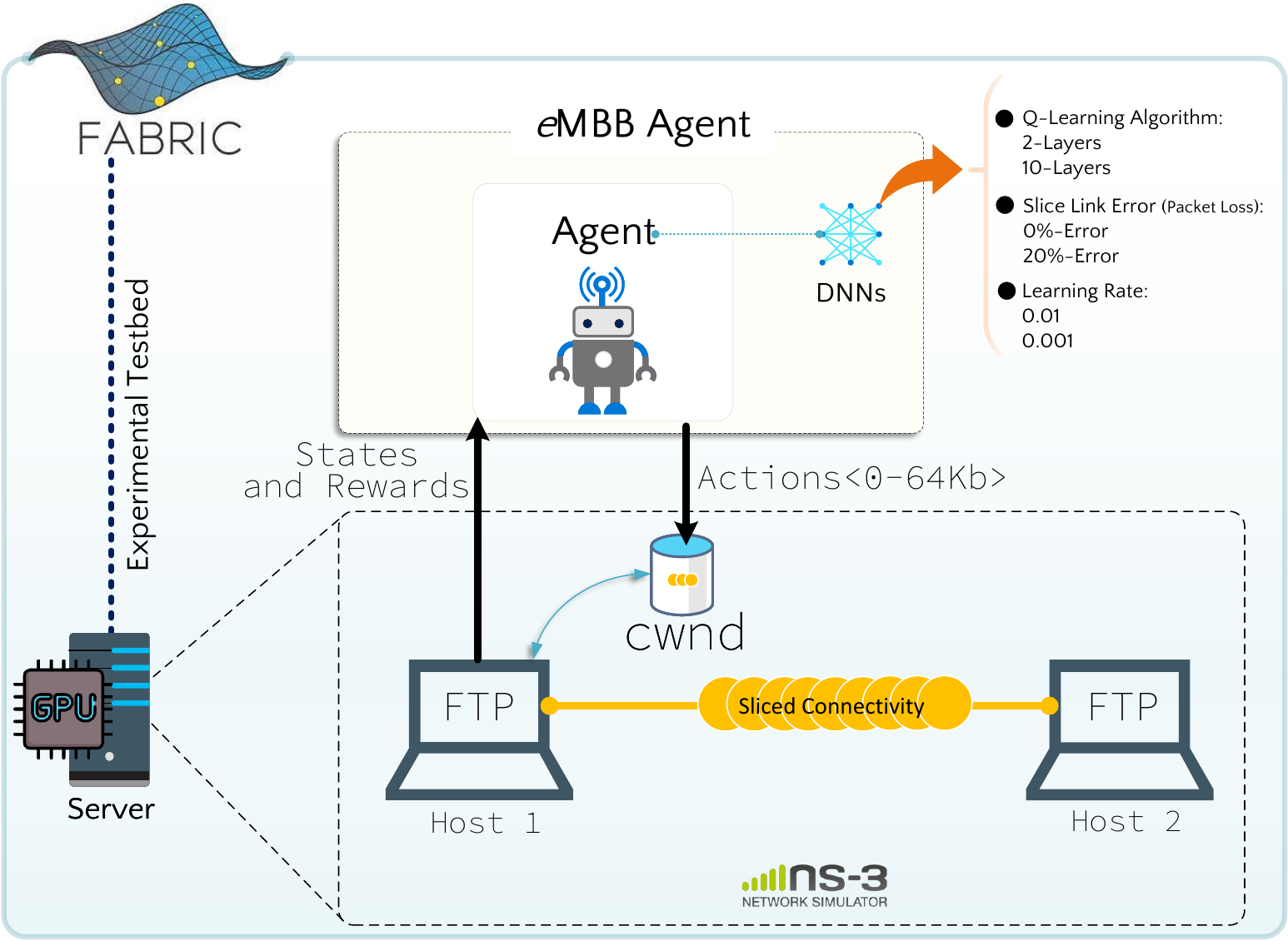}
  \caption{Proposed Evaluation Method.}
  \label{fig:method}
\end{figure}

This study examines the influence of various factors, including the number of layers in the  \ac{DQN} model, percentage of channel error, and learning rate, on the convergence time and data transmission rate (throughput) between two applications. The study employs metrics such as congestion window size, packet size, total number of bytes sent, average, total number of recognized segments, and network throughput. The aim is to optimize the current throughput on a link generated by the NS3 simulator, focusing on the impact of the \texttt{eMBB-Agent} on the  search space, as shown in Fig. ~\ref{fig:method}.

\begin{figure}[htbp]
  \centering
  \includegraphics[width=0.55\textwidth]{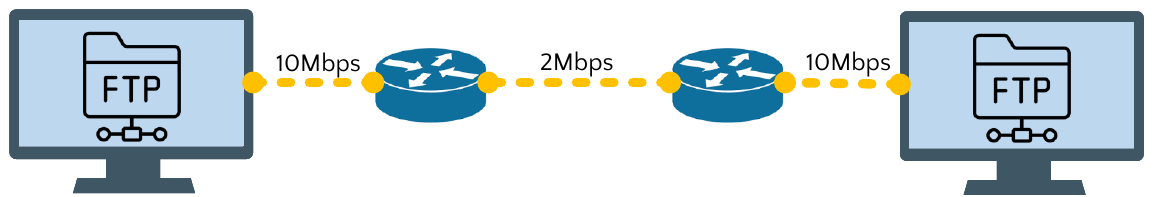}
  \caption{Experiment Topology.}
  \label{fig:experimental_topology}
\end{figure}

We used \ac{NS3} to create a network topology and simulate the transmission of packets between two \ac{FTP} applications, as shown in Fig. ~\ref{fig:method}. We set the bandwidth between the hots to 10 Mbps and 2 Mbps between the routers to induce congestion, as illustrated in Fig. ~\ref{fig:experimental_topology}. We used configurations \textit{NN-2} containing two hidden layers, \textit{NN-4} with four hidden layers, and \textit{NN-8} with eight hidden layers.

Through partial factorial combination, that is, we take combinations two-by-two and carry out experiments on the levels of variations to verify the influence of these variations. Each combination was run 10 times to generate a statistical sample. We measured the error rate in packets in scenarios of $0\%$ and $20\%$, and varied the learning rate hyperparameter to 0.01 and 0.001. The parameters of the combinations performed are listed in Table~\ref{tab:experimental_combinations}.

\begin{table}[htbp]
\centering
\scriptsize
\caption{Factorial-partial Experimentation Combinations.}
\label{tab:experimental_combinations}
\begin{tabular}{ccccc}
\hline
\multicolumn{1}{c}{\textbf{Factor}} & \multicolumn{4}{c}{\textbf{Levels}}                 \\ \hline
\textbf{\# Layers DQN Algorithm}    & 2           & 4          & \multicolumn{2}{c}{8}    \\
\textbf{Learning Rate}              & \multicolumn{2}{l}{0.01} & \multicolumn{2}{l}{0.001} \\
\textbf{Network Error Rate}            & \multicolumn{2}{l}{0\%}  & \multicolumn{2}{l}{20\%} \\ \hline
\end{tabular}
\end{table}

To train the \ac{DQN} models and their variations, we used RTX 4060\textit{ti} 16 Gb GPU hardware with an Intel(R) Core (TM) i5-4430 CPU @ 3.00GHz with 32 GB of \ ac{RAM}.

\section{Results and Discussion}\label{sec:results_and_discussion}

The central objective of the experiments was to analyze the behavior of the network in response to different configurations of \ac{NNs} by using the \ac{DQN} algorithm with the NS3-GYM~\cite{Gawlowicz2018} tool. With this technology, it has become possible to combine \ac{RL} algorithms and interventions in the control mechanisms of communication networks by using NS3. In this study we considered NS3-GYM online.

\begin{figure}[htbp]
\centering
\begin{subfigure}{.48\textwidth}
  \centering
  \includegraphics[width=\linewidth]{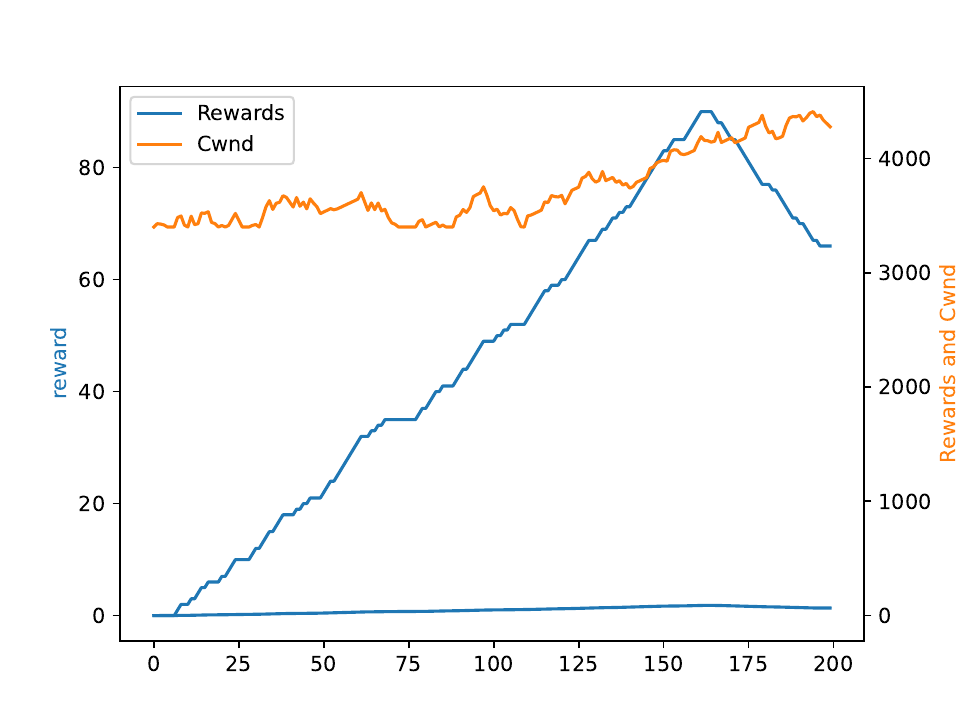}
  \caption{NN-2}
  \label{fig:flat}
\end{subfigure}%
\begin{subfigure}{.48\textwidth}
  \centering
  \includegraphics[width=\linewidth]{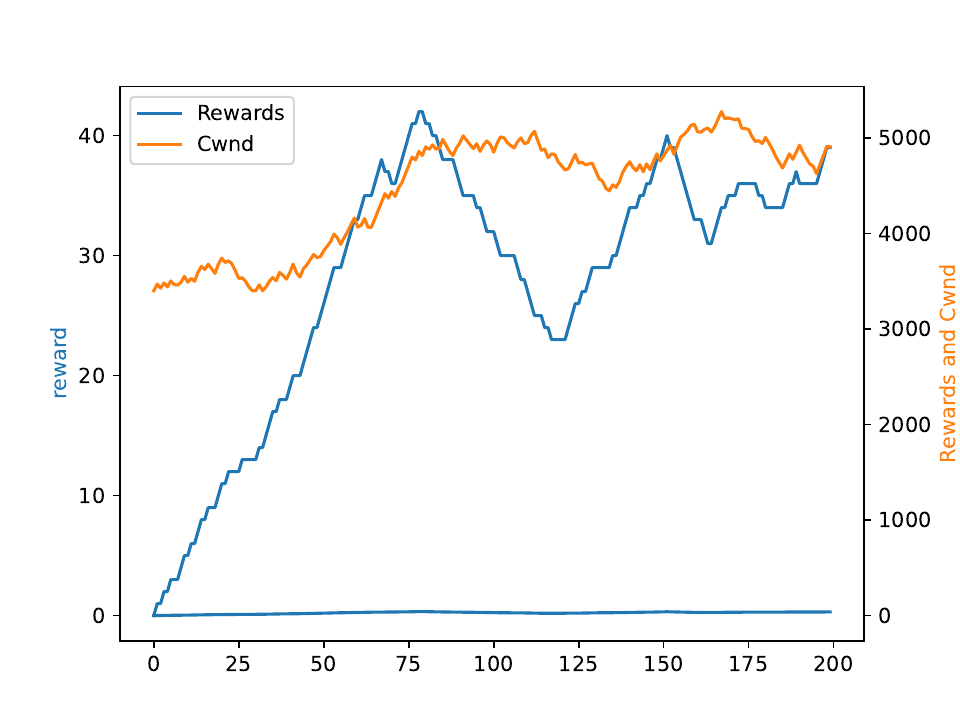}
  \caption{NN-4}
  \label{fig:NN}
\end{subfigure}

\begin{subfigure}{\textwidth}
  \centering
  \includegraphics[width=.48\linewidth]{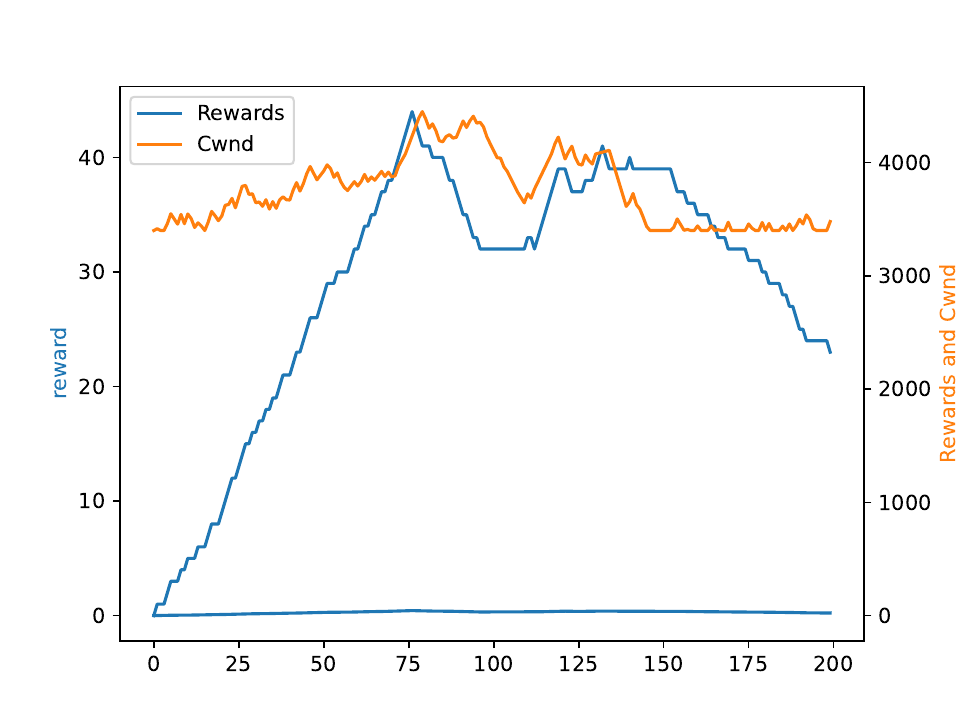}
  \caption{NN-8}
  \label{fig:DPM}
\end{subfigure}
\caption{Increasing \textit{cwnd} using different \ac{DQN} structures.}
\label{fig:congestion_window_increasing_over_epochs}
\end{figure}

During the simulation, a progressive increase in the congestion window was observed, which was directly related to a significant increase in the rewards obtained by the \texttt{eMBB-Agent}. According to Fig.~\ref{fig:congestion_window_increasing_over_epochs}, advantageous network flow rates were identified with the implementation of the \textit{\ac{NN}-2} model containing two layers. In detail, Fig.~\ref{fig:flat}, \ref{fig:NN} and \ref{fig:DPM} provide a visual representation of the algorithm over 200 \textit{steps}. Thus, it is possible to verify that the larger the size of the \textit{cwnd} variable in a smaller number of epochs, the faster \texttt{eMBB-Agent} increases the communication throughput.

\begin{figure}[htbp]
\centering
\begin{subfigure}{.5\textwidth}
  \centering
  \includegraphics[width=.99\linewidth]{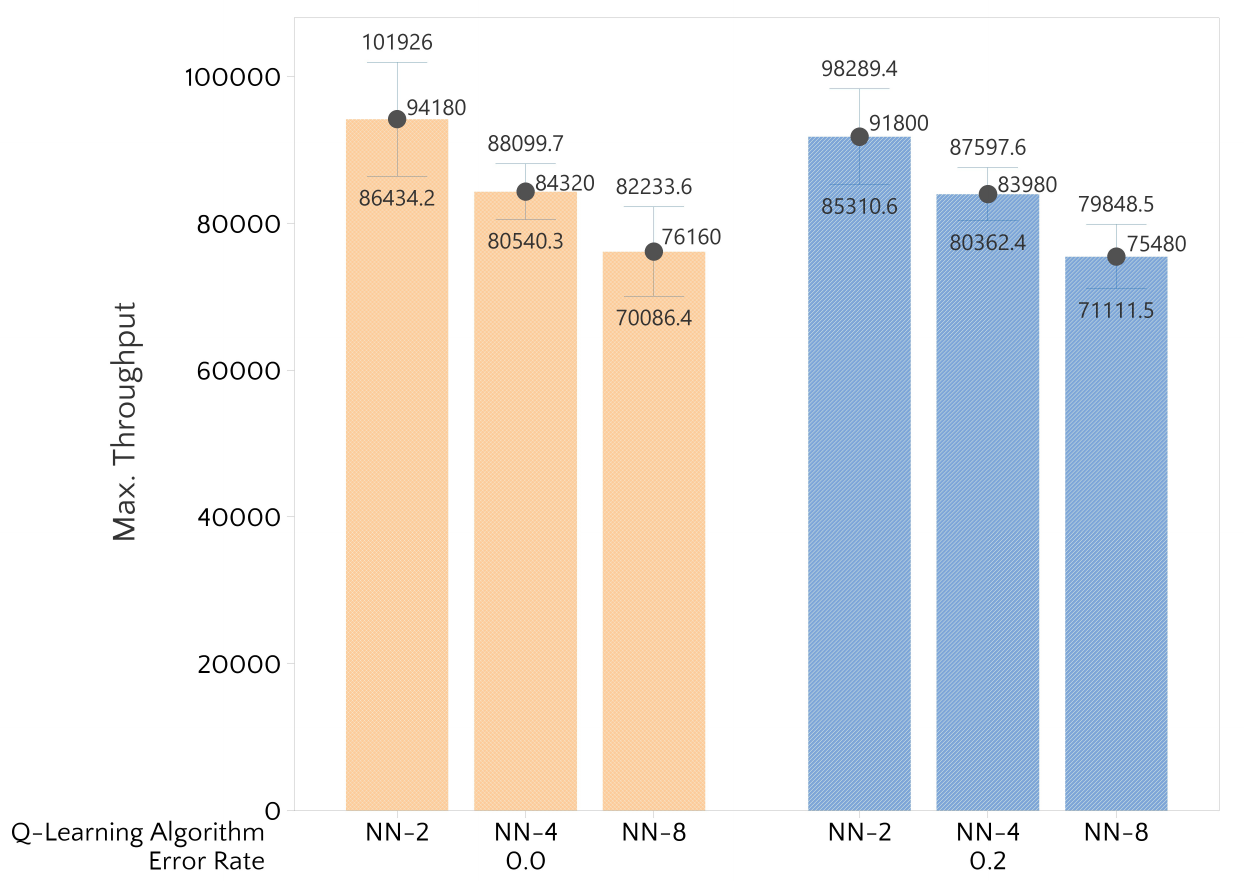}
  \caption{Maximum throughput achieved by different \textit{\ac{NN}} configurations.}
  \label{fig:througput_error_rate_learning_rate}
\end{subfigure}%
\begin{subfigure}{.5\textwidth}
  \centering
  \includegraphics[width=.99\linewidth]{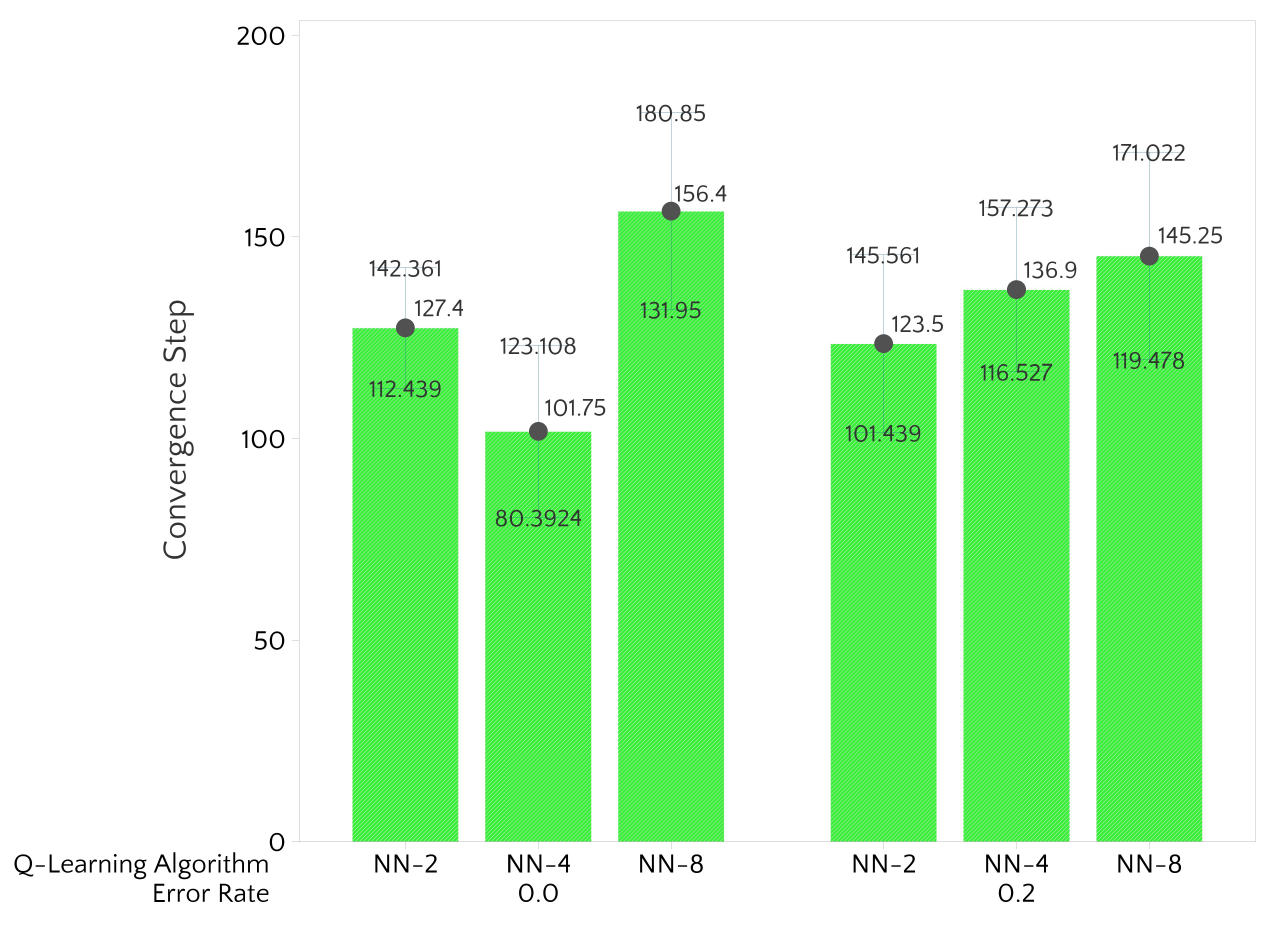}
  \caption{Convergence step required by different \textit{\ac{NN}} configurations.}
  \label{fig:convergence_steop_error_rate}
\end{subfigure}
\caption{Analyses of a physical link with ($20\%$) and without ($0\%$) errors.}
\label{fig:test}
\end{figure}

The \textit{\ac{NN}} configuration, shown in Fig. ~\ref{fig:flat}, \ref{fig:NN} and \ref{fig:DPM} exhibit a progressive increase in throughput with an increasing congestion window \textit{cwnd} based on decisions and receiving rewards for correct choices. Three models (\textit{\ac{NN}-2}, \textit{\ac{NN}-4}, and \textit{\ac{NN}-8}) were analyzed, and \textit{\ac {NN}-4} presented a lower average flow rate considering \textit{\ac{NN}-2} and \textit{\ac{NN}-8}. Subsequently, \textit{\ac{NN}-8} manifested the second-worst throughput. This behavior is attributed to the time required to train a \ac{DQN} with more layers, given the time sensitivity in the experimental scenario.

Fig.~\ref{fig:througput_error_rate_learning_rate} highlights the performance of the \textit{Q-Learning} algorithm in three configurations: \textit{\ac{NN}-2}, \textit{\ac{NN}-4} and \textit{\ac{NN}-8}, representing the average throughput of the network slice in an error-free and with error in slicing. \textit{\ac{NN}-2}, with an error-free demonstrates the best performance, followed by \textit{\ac{NN}-4}, while \textit{\ac{NN}-8} displays lower performance, with a lower Average Network Throughput. We associate this with its complexity, which requires more computational resources owing to its deep neural network structure.

Fig.~\ref{fig:convergence_steop_error_rate} shows the convergence times of the \ac{DQN} algorithm for three different architectures: \textit{\ac{NN}-2}, \textit{\ac{NN}-4 }, and \textit{\ac{NN}-8}. As can be seen, in an error-free channel, algorithm \textit{\ac{NN}-4} exhibits the best performance, with a convergence time $20.09\%$ lower than \textit{\ac{NN}- 2} and $32.99\%$ lower than \textit{\ac{NN}-8}, respectively.

Alternatively, in a channel with $20\%$ error induction, \textit{\ac{NN}-2} reaches convergence, that is, fullness in the second variable faster. Thus, the convergence time for \textit{\ac{NN}-2} was $10.34\%$ less than that of \textit{\ac{NN}-4} and $15.00\%$ less than that of \textit{\ac{NN}- 8}. We associate this better performance of the simpler \textit{\ac{NN}} with better training time.

\begin{table}[!htbp]
\caption{Influence of \textit{network error rate} and \ac{DQN} structure on Network Throughput.}
  \centering
  \scriptsize
  \resizebox{\textwidth}{!}{%
    \begin{tabular}{|c|c|c|c|c|c|}
    \hline
    \textbf{Term} & \textbf{Influence} & \textbf{Coefficient} & \textbf{Standard Error} & \textbf{T-\textit{Value}} & \textbf{P-\textit{Value}} \\
    \hline
    Constant & & 84320 & 1075 & 78.41 & 0.000 \\
    Network Error Rate & -1133 & -567 & 1075 & -0.53 & 0.599 \\
    \ac{DQN} Algorithm & -17170 & -8585 & 1317 & -6.52 & 0.000 \\
    Network Error Rate $\times$ \ac{DQN} Algorithm & 850 & 425 & 1317 & 0.32 & 0.748 \\
    \hline
    \end{tabular}%
  }
    \label{tab:ResultadosRegressao1}
  \end{table}

\begin{table}[!htbp]
    \caption{Influence of \textit{network error rate} and \textit{learning rate} on Network Throughput.}
  \centering
  \scriptsize
  \resizebox{\textwidth}{!}{%
    \begin{tabular}{|c|c|c|c|c|c|}
    \hline
    \textbf{Term} & \textbf{Influence} & \textbf{Coefficient} & \textbf{Standard Error} & \textbf{T-\textit{Value}} & \textbf{P-\textit{Value}} \\
    \hline
    Constant & & 84320 & 1075 & 78.41 & 0.000 \\
    Network Error Rate & -1133 & -567 & 1075 & -0.53 & 0.599 \\
    Learning Rate & 907 & 453 & 1075 & 0.42 & 0.678 \\
    Network Error Rate $\times$ Learning Rate & 680 & 340 & 1075 & 0.32 & 0.748 \\
    \hline
    \end{tabular}%
  }
  \label{tab:ResultadosRegressao2}
\end{table}

Finally, we investigate the effect of the error rate, learning rate, and \ac{RL} algorithm on network throughput through regression. Tables~\ref{tab:ResultadosRegressao1} and~\ref{tab:ResultadosRegressao2} present the variables considered, their estimated impacts, associated coefficients, standard errors, and T and P-\textit{values}, providing information about the relationships between Algorithm \textit{DQN }, error rate, learning rate, and network throughput.

\section{Concluding Remark}\label{sec:concluding_remark}

The analysis of congestion algorithms with various combinations of artificial \textit{\ac{NN}} demonstrated an inverse correlation between the number of layers and the efficiency of optimizing the flow in communication networks, as indicated by the data from linear regression analysis. This suggests that an increase in network complexity leads to a decrease in network throughput. The study found that neither the network error rate nor the learning rate had a statistically significant effect on the network throughput.

This study struggled with some constraints, one of which was that the analysis was carried out in a simulated environment using Network Simulation Library 3 (NS3). To overcome this limitation, we suggest that future research explore the functionality of \texttt{\ac{eMBB}-Agent} in a real-world setting, in which variables can be manipulated to assess its impact in more complex and dynamic situations. This provided a more accurate and comprehensive understanding of the practical implications of the findings across various operational scenarios. In addition, we suggest the optimization of additional parameters and evaluation of latency and reliability to further improve network performance.

\section*{Acknowledgments}

We acknowledge the financial support of the FAPESP MCTIC/CGI Research project 2018/23097-3 - SFI2 - Slicing Future Internet Infrastructures. This study was financed in part by the Coordenação de Aperfeiçoamento de Pessoal de Nível Superior - Brasil (CAPES) - Finance Code 001.

\bibliographystyle{sbc}
\bibliography{Template_SBC/references}

\end{document}